\documentclass[11pt]{article}

\usepackage[preprint]{acl}

\usepackage{times}
\usepackage{latexsym}

\usepackage[T1]{fontenc}

\usepackage[utf8]{inputenc}

\usepackage{microtype}

\usepackage{inconsolata}

\usepackage{graphicx}

\usepackage{hyperref}
\usepackage{url}
\usepackage{paralist}
\usepackage{mathtools,amsthm,amssymb}
\usepackage{multirow}
\usepackage{booktabs}
\usepackage{xspace}
\usepackage{pifont}
\usepackage{tabularx}
\usepackage{caption}
\usepackage{array}
\newcolumntype{C}[1]{>{\centering\arraybackslash}m{#1}}
\newcolumntype{L}[1]{>{\arraybackslash}m{#1}}
\newcolumntype{P}[1]{>{\raggedright\arraybackslash}p{#1}}

\usepackage{algorithm}
\usepackage{algpseudocode}
\usepackage{setspace}
\usepackage{xcolor,colortbl}
\usepackage{enumitem}
\usepackage{wrapfig}
\usepackage{makecell}
\usepackage{dsfont}

\newcommand{\ours}{\textsc{ToolQP}\xspace}

\newcommand{\1}[1]{\mathds{1}\left[#1\right]}

\newcommand{\Append}{\textsc{Append}}
\newcommand{\Extend}{\textsc{Extend}}

\algnewcommand\algorithmicforeach{\textbf{for each}}
\algdef{S}[FOR]{ForEach}[1]{\algorithmicforeach\ #1\ \algorithmicdo}

\title{Beyond Single-Shot: Multi-step Tool Retrieval via Query Planning}

\author{
  Wei Fang \and James Glass
\\
  Massachusetts Institute of Technology, Cambridge MA, USA
\\
\texttt{\hypersetup{urlcolor=black}\{\href{mailto:weifang@mit.edu}{weifang},\href{mailto:glass@mit.edu}{glass}\}@mit.edu}
}

\begin{document}
\maketitle
\begin{abstract}
LLM agents operating over massive, dynamic tool libraries rely on effective retrieval, yet standard single-shot dense retrievers struggle with complex requests. 
These failures primarily stem from the disconnect between abstract user goals and technical documentation, and the limited capacity of fixed-size embeddings to model combinatorial tool compositions. 
To address these challenges, we propose \ours, a lightweight framework that models retrieval as iterative query planning. 
Instead of single-shot matching, \ours decomposes instructions into sub-tasks and dynamically generates queries to interact with the retriever, effectively bridging the semantic gap by targeting the specific sub-tasks required for composition. 
We train \ours using synthetic query trajectories followed by optimization via Reinforcement Learning with Verifiable Rewards (RLVR). 
Experiments demonstrate that \ours achieves state-of-the-art performance, exhibiting superior zero-shot generalization, robustness across diverse retrievers, and significant improvements in downstream agentic execution.

\end{abstract}

\section{Introduction}
Large language models (LLMs) have evolved from simple text generation into integration within agentic frameworks that allow them to solve a variety of complex tasks such as math, reasoning, and coding, by interacting with external environments~\citep{mialon2023augmented,yao2023react}. 
Integral to this paradigm shift is the ability to use tools, namely APIs, databases, and software tools, to extend the models' capabilities beyond their parametric knowledge~\citep{qin2024toollearningfoundationmodels}.
As agentic workflows are being developed, the scale of these tool libraries are expanding rapidly, moving from dozens of hand-picked functions to massive, dynamic repositories containing tens of thousands of APIs.
In these scenarios, it is computationally infeasible to fit the entire tool context, including documentation, tool-specific instructions, and in-context tool demonstrations, into the LLM’s context window.
Consequently, tool retrieval has been fundamental to the design of practical frameworks, retrieving relevant tools from the toolset as an initial step~\citep{li-etal-2023-api}.

While recent work has adapted information retrieval (IR) techniques with ad-hoc tool-use datasets~\citep{qu_towards_2024,xu-etal-2024-enhancing-tool} to enhance tool retrieval, they along with approaches that perform well on conventional IR benchmarks are shown to exhibit poor performance on a wide variety of tool-use tasks and tools~\citep{shi-etal-2025-retrieval}. 
These existing approaches typically employ dense embeddings with a standard single-shot retrieval step, and while they may be effective for simple, direct queries, they often fail significantly when applied to complex, compositional tasks. 

We identify three fundamental challenges that stem from applying these single-shot retrieval paradigms to dynamic agentic workflows. 
First, semantic misalignment creates a critical disconnect between the high-level vocabulary of user intent and the technical specificity of tool schemas. 
For instance, a user may request to "make this audio recording high quality," while a relevant tool \texttt{scipy.signal.lfilter(b, a, x)} may be defined strictly by mathematical parameters like \texttt{b} (numerator) and \texttt{a} (denominator) along with technical descriptions such as ``Filter data along one-dimension with an IIR or FIR filter.''. 
Standard dense retrievers fail to bridge the gap between the subjective goal ("high quality") and the implementation-level terminology (\texttt{lfilter}), and this is exacerbated by the heterogeneous nature of large-scale tool libraries, where documentation styles vary from verbose descriptions to raw, schema-heavy protocols~\citep{qin2024toolllm,shi-etal-2025-retrieval}.
Furthermore, real-world tasks are inherently compositional, often requiring the simultaneous application of multiple distinct tools; for example, retrieving both a \texttt{WeatherAPI} and a \texttt{StockMarketDB} to ``analyze how rain affects retail sales.''
However, a single fixed-dimensional vector lacks the capacity to encode the combinatorial diversity of multiple disparate tools~\citep{weller_theoretical_2025}, a limitation that is amplified as tool libraries scale. 
Finally, current single-shot methods lack interactive toolset awareness.
They treat the repository as a static database and cannot handle internal constraints or changes. 
In contrast, interacting with the environment provides critical feedback on inter-tool dependencies~\citep{xu-etal-2024-enhancing-tool}, for example discovering that a \texttt{forecasting\_tool} requires a specific \texttt{region\_id} from a lookup utility, and allows the system to adapt to modifications within the toolset.

To address these limitations, we propose the Tool Query Planner (\ours), a framework that formulates retrieval as an iterative planning process rather than a static single-shot semantic matching task. \ours decomposes complex user requests into a logical sequence of high-level sub-tasks, interactively retrieving relevant tools for each step through a unified and light-weight model designed to interface with any existing retrieval system.
This approach effectively bridges semantic gaps by inferring functional utility from abstract goals, circumvents compositional bottlenecks by retrieving conceptually-similar tools step-by-step rather than compressing them into a single vector, and resolves inter-tool dependencies and toolset modifications via dynamic environment feedback.
Our design is inherently modular and generalizable, functioning as a complementary layer atop standard retrievers without requiring architectural changes to the underlying index or the downstream reasoning LLM. 
Furthermore, the explicit planning trajectory generated during retrieval could serve as valuable context for the downstream agent to ground its execution.
Extensive experiments across a wide variety of tool-use tasks demonstrate that \ours significantly improves both retrieval accuracy and downstream execution success rates compared to state-of-the-art baselines. 
Overall, our contributions are summarized as follows:
\begin{itemize}
    \item We propose \ours, a novel framework that fundamentally shifts tool retrieval from a static similarity matching task to a dynamic planning process. By reframing the problem, we enable the resolution of complex, compositional queries and facilitate the discovery of inter-tool dependencies that single-shot dense retrievers are inherently limited in addressing.
    \item We design \ours as a modular, lightweight layer that integrates seamlessly with existing dense retrievers and downstream LLMs. Our approach leverages interactive feedback to adapt to heterogeneous tool documentation styles and diverse retrieval environments without requiring architectural modifications or fine-tuning of the underlying system.
    \item We demonstrate through extensive experiments on a diverse set of tool-use benchmarks that \ours significantly outperforms state-of-the-art baselines. Our results show consistent improvements in both retrieval performance and downstream execution success rates, particularly in scenarios characterized by high compositional complexity and abstract user intent.
\end{itemize}

\section{Related Work}

\begin{figure*}[ht!]
    \centering
    \includegraphics[width=0.98\linewidth]{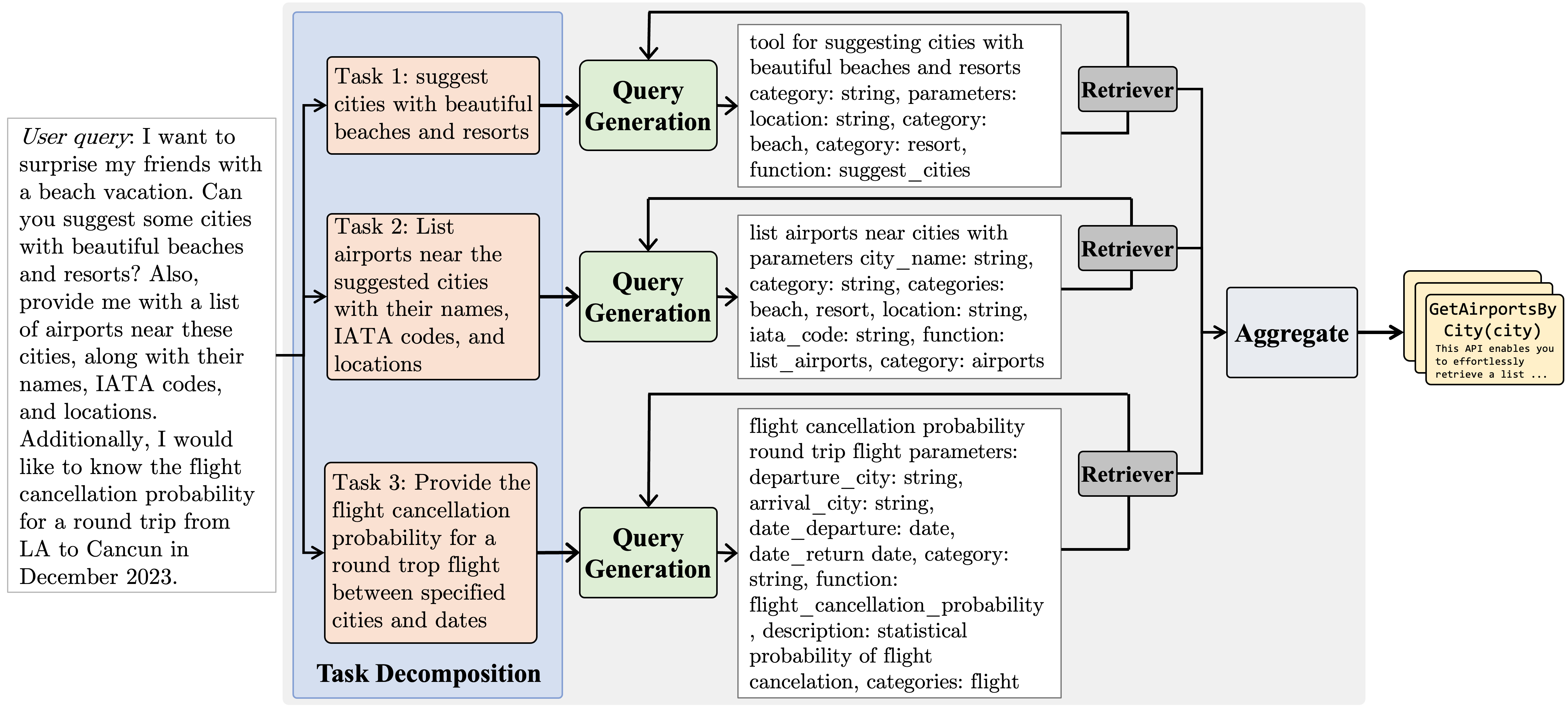}
    \caption{Overview of the \ours framework. The Planner decomposes a complex user query (e.g., travel planning) into sequential sub-tasks. For each sub-task, it interactively generates queries, processes feedback from the dense retriever, and self-corrects if necessary, before aggregating the final set of relevant tools.}
    \label{fig:framework}
\end{figure*}
\paragraph{Tool Learning and Retrieval.}
LLMs are increasingly employed in agentic frameworks that enable tool use for solving complex tasks~\citep{gupta2023visual,mialon2023augmented,suris2023vipergpt,team2023gemini,wu2023visual,cai2024large,qin2024toollearningfoundationmodels,zhang-etal-2024-natural}.
Conventional approaches include post-training fine-tuning~\citep{parisi2022talm,thoppilan2022lamda,patil2023gorilla,schick2023toolformer,dubey2024llama,yang2024gpt4tools,liu2025toolace,lin2025robust,yang2025qwen3}, or in-context learning with meta-prompts for zero-shot tool usage~\citep{lu2023chameleon,shen2023hugginggpt,song2023restgpt,qin2024toolllm,zhuang2024toolchain}. 
However, scaling to large toolsets (e.g., 52k+ in RapidAPI) is challenging due to limited context windows and performance degradation from long contexts~\citep{liu_apigen_2024,qu_towards_2024}, an issue exacerbated when including necessary instructions and demonstrations~\citep{hsieh2023tool,xu2023tool}.
Furthermore, frequent updates to the toolset make retraining cost-prohibitive, necessitating zero-shot approaches~\citep{fang-etal-2025-play2prompt,qu2025from}.
Contemporary tool-use frameworks address this via semantic retrievers, utilizing either conventional dense embeddings ill-suited for tools~\citep{shi-etal-2025-retrieval} or task-specific models lacking generalizability~\citep{gao_confucius_2024,kong-etal-2024-tptu,qin2024toolllm,wang2025toolgen}.
Tool retrieval is arguably more challenging than conventional IR due to multi-tool composition~\citep{qu_towards_2024} and the semantic gap between user intent and technical documentations~\citep{chen-etal-2024-invoke}, and \ours is explicitly designed to address these challenges.

\paragraph{Generative Modeling for Retrieval.}
Document expansion, which appends generated queries to documents, and query expansion, which augments queries with relevant content, are established IR techniques that are simple yet effective~\citep{10.1145/321033.321035,10.1145/160688.160713,10.1145/243199.243202,10.1145/312624.312645}. 
Pseudo-relevance feedback similarly expands queries using top-ranked documents from a first-pass retrieval~\citep{1570291225586316544,croft_using_1979,10.1145/383952.383972,10.1145/3471158.3472250,10.1145/3459637.3482124,10.1145/3570724}.
Recently, LLMs have advanced generative document expansion for IR and QA~\citep{dai2019context,nogueira2019document,10.1145/3404835.3463098,lewis-etal-2021-paq}, and query expansion via hypothetical document generation~\citep{gao-etal-2023-precise,jagerman_query_2023,10.1145/3539618.3591992,shen2023large,wang-etal-2023-query2doc} or expansion term selection~\citep{mao-etal-2021-generation,chuang-etal-2023-expand}.
Alternatively, LLMs facilitate query rewriting--reformulating the user query entirely--and retrieval scoring and re-ranking~\citep{10.1145/3397271.3401323,sachan-etal-2022-improving,ma-etal-2023-query,mao-etal-2023-large,sun-etal-2023-chatgpt,ye-etal-2023-enhancing,fang-etal-2024-joint,feng-etal-2024-synergistic}.
Recent studies adapt these methods for tool retrieval via few-shot prompting~\citep{chen-etal-2024-invoke}, or by incorporating them within the retriever fine-tuning process~\citep{xu-etal-2024-enhancing-tool}. 
Contemporaneous work further explores training with large-scale expansion augmentation~\citep{sengupta2025tooldreamer,lu2025tools}.
In comparison, \ours learns robust query planning while being very lightweight, outperforming without prompting large models or requiring retriever fine-tuning.

\section{\ours: Query Planning}
\subsection{The Modeling Framework}
We assume access to a tool retriever $\mathcal{E}$ that receives a user query $q$ and returns a ranked list of tools from the tool set $\mathcal{D}$.
A typical single-shot retrieval directly embeds $q$ and all tools $d\in\mathcal{D}$ using the retriever $\mathcal{E}$ and selects the tools with highest similarity score as relevant tools.
However, as previously discussed, this single-shot paradigm struggles to capture compositional intent and lacks interactivity required to resolve dynamic changes to the toolset or tool-specific dependencies.
To address these limitations, we propose \ours, a query planning framework that casts tool retrieval as a sequential decision-making process.
Rather than treating the retriever as a static index, $\mathcal{E}$ is treated as a dynamic environment for which \ours can interact with and explore the tool space.
The \ours framework consists of three stages: \textit{planning}, \textit{query generation}, and \textit{aggregation}, as illustrated in Fig.~\ref{fig:framework}.

\paragraph{Planning by Task Decomposition.} The primary challenge in tool retrieval is the semantic misalignment between high-level user intents and low-level tool representations.
Directly querying the retriever with a complex instruction $q$ often leads to suboptimal performance since the necessary tool keywords are usually absent from the user's phrasing.
To bridge this gap, \ours first generates a natural language plan $\mathcal{P}$, and then decomposes the user query $q$ into a logical sequence of sub-tasks $\{s_n\}$.
Unlike prior methods that rely on costly, prompt-engineered calls to large models to identify user intents, \ours is a lightweight and modular solution that learns this decomposition capability directly and efficiently.

\paragraph{Interactive Query Generation.} Guided by the generated plan $\mathcal{P}$, \ours then targets each sub-task by interactively generating a sequence of search queries $\{q_t\}$.
At each step $t$, the model observes the retrieval feedback $O_t=\mathcal{E}(q_t; \mathcal{D})$ before generating the next query $q_{t+1}$.
This feedback loop addresses the limitations of single-shot retrieval by allowing the model to dynamically adjust its strategy based on the retrieval feedback.
For example, if an initial query retrieves a relevant tool that requires a specific input argument, the model can generate subsequent queries to target that prerequisite tool.
This iterative process continues until the model determines the sub-tasks in $\mathcal{P}$ have been sufficiently covered, resulting in a trajectory of query-retrieval pairs $\{(q_t, O_t)\}$.

\paragraph{Retrieval Aggregation.} The final stage is to aggregate results from the query-retrieval trajectory into a single ranked list that can be used by a downstream LLM.
While previous work have explored reciprocal rank fusion (RRF) or engineered complex rank-fusion algorithms to integrate multiple retrieval rankings, we find these methods unsuitable or unnecessary for our planning-based framework.
Since the model may generate a varying number of queries for different sub-tasks, prior methods tend to bias the final ranking toward sub-tasks that require more query attempts.
Therefore, we employ a simple yet robust \emph{peak-rank} aggregation strategy: for every unique tool $d\in\bigcup O_t$, we assign its final rank based strictly on the highest rank it achieved across any single retrieval attempt.
This effectively balances the retrieval results across each sub-tasks.

\subsection{Data Generation \& SFT}
To train a \ours model, it requires supervision for the modeling outputs we described earlier: the task decomposition plan $\mathcal{P}$, and the subsequent search query trajectory $\{(q_t, O_t)\}$.
However, standard tool retrieval datasets typically provide only the user query $q$ and the final set of ground-truth tools $\mathcal{T}^\star$. 
While ~\citet{shi-etal-2025-retrieval} previously paired each instance with a high-level instruction, which we can use as target for $\mathcal{P}$, we still lack the query trajectories required for training.
We thus design a data synthesis pipeline to generate effective query trajectories from data compiled by ~\citet{shi-etal-2025-retrieval} using a teacher model.
The process, illustrated in Alg.~\ref{alg:datagen}, involves three stages, which we describe below.

\paragraph{Plan Alignment.} First, we must ground the high-level natural language plan $\mathcal{P}$ in the ground-truth tools $\mathcal{T}^\star$.
We prompt a teacher model to parse $\mathcal{P}$ into a sequence of discrete sub-tasks and assign the corresponding subset of target tools $t\in\mathcal{T}^\star$ to each sub-task.
This results in a reasonable task decomposition and also a clear mapping between each sub-task and the tools required to complete them.

\paragraph{Query Generation.} Next, we generate the search queries necessary to retrieve the assigned tools for each sub-task.
To prevent generating trivial queries such as using the exact target tool names, we use the teacher model to simulate the query generation process, without conditioning on the specific target tool names; specifically, we prompt the teacher model to generate $N$ candidate queries for a sub-task based only on the sub-task description.
This constraint forces the generation of queries that describe the desired functionality of the tool rather than memorized identifiers.
However, for tools that are difficult to retrieve with generic descriptions, we adopt a curriculum learning approach: if the initial query candidates fail to retrieve the target tools (as determined in the verification step below), we iteratively re-prompt the teacher with more target information until a valid query is found.

\paragraph{Query Verification and Trajectory Construction.} For each step, we input all $N$ candidate queries to the retriever $\mathcal{E}$ to obtain the ranks of the target tools, and select the query candidate that achieves the highest recall and is higher than a set threshold rank $r_\tau$ to be the valid query.
To construct a full trajectory of $\{(q_t, O_t)\}$, we gather all valid queries $q_t$ for all sub-tasks of the user query $q$. 
Additionally, we include query sequences that include \emph{failed} attempts that yielded low ranks followed by a subsequent \emph{successful} query to simulate an interactive trial-and-error process, providing the model with explicit training examples of self-correction.
The resulting synthetic data trajectory, comprising the high-level plan $P$, sub-tasks $\{s_n\}$, and query-retrieval feedback sequence $\{(q_t, O_t)\}$, is used to train the model via standard supervised fine-tuning with maximum likelihood and teacher forcing.
\begin{algorithm}[t!]
\caption{\sc TrajectoryGeneration}
\label{alg:datagen}
\begin{algorithmic}[1]
\small
\Require $q$: user query, $\mathcal{P}$: natural language plan, target tools $\mathcal{T}^\star$, teacher model $\mathcal{M}$
\State $\{(s_n,\mathcal{T}^\star_n)\}_{n=1}^{K}\gets\textsc{Parse}_\mathcal{M}(q,\mathcal{P})$\Comment{Parse sub-tasks}
\State $\mathcal{H}\gets[\mathcal{P},\,s_1,\,\dots,\,s_K]$
\For{$n\gets 1$ \textbf{to} $K$}
    \State $\mathcal{A}\gets[\ ]$
    \While{$\textsc{AvgRank}(\mathcal{T}_n^\star,O_t)>r_\tau$}
        \State $c\gets\textsc{AddMoreInfo}(c,\mathcal{T}^\star_n)$\Comment{$c$: context}
        \State $Q_{\text{Cand}}\gets\{\textsc{Generate}_{\mathcal M}(q,s_{1:n-1},c)\}_{i=1}^{N}$
        \State $\mathcal{O}_{\text{Cand}}\gets\{\textsc{Retrieve}(Q_{\text{Cand}}^{(i)})\}_{i=1}^{N}$
        \State $i^\star\gets \operatorname*{Best}_{i\in\{1,\dots,N\}}\;
            \textsc{AvgRank}(\mathcal{T}_n^\star,\mathcal{O}_{\text{Cand}}^{(i)})$
        \State $q_t, O_t\gets Q_{\text{Cand}}^{(i^\star)}, \mathcal{O}_{\text{Cand}}^{(i^\star)}$
        \State $\Append(\mathcal{A},(q_t,O_t))$
    \EndWhile
    \If{$\textsc{Bern}(p)$} \Comment{Keep \textit{failed} attempts with prob. $p$}
        \State $\Extend(\mathcal{H},\mathcal{A})$
    \Else
        \State $\Append(\mathcal{H},(q_t,O_t))$
    \EndIf
\EndFor
\State \Return $\mathcal{H}$
\end{algorithmic}
\end{algorithm}

\subsection{Training -- RLVR}
While SFT provides a strong baseline capability by distilling the teacher's trajectories, to enable the model to explore the search space and discover query strategies that maximize retrieval performance, we further train the model using Reinforcement Learning with Verifiable Rewards (RLVR)~\citep{lambert2024tulu,guo2025deepseek,yue2025does}.
Specifically, we employ Group Relative Policy Optimization (GRPO)~\citep{shao2024deepseekmath}, which eliminates the need for a value network and leverages the deterministic nature of our retrieval environment.
The training objective optimizes an overall reward $\mathcal{R} = \beta_1 \mathcal{R}_{\text{retrieval}} + \beta_2 \mathcal{R}_{\text{format}} + \beta_3 \mathcal{R}_{\text{plan}}$.
The primary reward, $\mathcal{R}_{\text{retrieval}}$, is a sequence-level reward calculated via the nDCG@K and Recall@K of the final aggregated tool list against the ground truth, which encourages the model to optimize global coverage rather than individual steps.
$\mathcal{R}_{\text{format}}$ is used to enforce formatting validity, while  $\mathcal{R}_{\text{plan}}$ serves as a regularizer that computes the semantic similarity between the generated plan and the original high-level plan $\mathcal{P}$, preventing the model from deviating from the user's intent while exploring valid decompositions.

\section{Experiments}

\subsection{Tool Retrieval}

\paragraph{Benchmark and Setup.} We evaluate \ours on ToolRet~\cite{shi-etal-2025-retrieval}, a comprehensive benchmark comprising 35 widely-used tool-calling datasets categorized into \emph{Web}, \emph{Code}, and \emph{Custom} domains, with an overall toolset of 44k tools.
For training \ours, we utilize a subset of the training split compiled by \citet{shi-etal-2025-retrieval}, which is sourced from the ToolBench, ToolACE, and APIGen training sets.
Since these training sources fall under the Web category, we treat the corresponding test sets as \emph{in-domain}, while the rest constitute a true \emph{zero-shot transfer} setting.
Following prior work, we report the standard IR metric Normalized Discounted Cumulative Gain (nDCG@K), and Completeness@K ($\1{R@K=1}$), with $K=10$ and macro-averaging across datasets within the same category.
For this setting, we use the retriever \texttt{gte-Qwen2-1.5B-instruct} \citep{li2023towards} (abbrev. \texttt{gte-Qwen}) as the base retriever for the baseline methods and \ours.
Additional details about the data and licenses can be found in Appx~\ref{appx:data}.
\begingroup
\setlength{\tabcolsep}{4pt}
\begin{table*}[t!]
\small
\centering
\begin{tabular}{L{3.4cm} >{\columncolor[gray]{0.9}}c>{\columncolor[gray]{0.9}}c  cc cc cc >{\columncolor[gray]{0.9}}c>{\columncolor[gray]{0.9}}c}
\toprule
\multirow{3}{*}[-1.2ex]{\bf Method} 
    & \multicolumn{2}{c}{\bf In-Domain} 
    & \multicolumn{8}{c}{\bf Zero-Shot Transfer} \\
    
\cmidrule(lr){2-3}
\cmidrule(lr){4-11}
& \multicolumn{2}{c}{\cellcolor[gray]{0.9}\bf Avg} 
& \multicolumn{2}{c}{\bf Web$^{\star}$} 
& \multicolumn{2}{c}{\bf Code} 
& \multicolumn{2}{c}{\bf Custom} 
& \multicolumn{2}{c}{\cellcolor[gray]{0.9}\bf Macro-Avg} \\

\cmidrule(lr){2-3}
\cmidrule(lr){4-11}

& N@10 & C@10 
& N@10 & C@10
& N@10 & C@10
& N@10 & C@10
& N@10 & C@10
\\
\midrule
Base Retriever (\texttt{gte-Qwen}) & 43.3 & 47.8 & 29.7 & 17.8 & 24.1 & 30.8 & 35.6 & 32.4 & 29.8 & 27.0 \\
\midrule
\multicolumn{11}{c}{\textit{Prompting} (\texttt{Qwen3-30B-A3B-Instruct})} \\
\midrule
Q2E/ZS & 44.6 & 49.0 & 30.8 & 19.6 & 26.5 & 34.0 & 38.7 & 35.7& 32.0& 29.8 \\
Q2D/ZS & 52.0 & 56.8 & 24.5 & 18.1  & 22.2 & 29.1  & 33.3 & 33.5&26.7& 26.9\\
HyDE/ZS & 47.9 & 52.3 & 19.7 & 17.0 & 13.6 & 18.8 & 26.4  & 29.7& 19.9& 21.8\\
D2Q & 51.5& 54.0 & 26.6  & 16.3 & 25.4 &  32.0 & 33.3  & 30.8 & 28.4&26.4\\
Re-Invoke & 51.5 & 56.8 & 28.1 & 18.5  & 26.0  & 33.5  & 36.8  & 31.7& 30.3& 27.9\\
\midrule
Q2E/PRF & 45.0 & 49.4 & 31.0 & 18.8 & 26.8 & 33.1 & 35.3 & 33.3 &  31.0& 28.4\\
Q2D/PRF & 46.9 & 50.2 & 30.5  & 18.9  & 26.1  & 31.1  & 37.0 & 33.7&31.2& 27.9\\
HyDE/PRF & 43.5 & 43.5 & 26.5  & 18.0  & 24.8  & 28.6 & 41.6 & 26.7&27.6&24.4\\
\midrule
\multicolumn{11}{c}{\it Re-ranking with cross-encoder} \\
\midrule
\texttt{bge-m3} & 49.4 & 52.5 & 32.7 & 18.5& 24.7 & 31.6 & 32.9& 30.2 & 30.1 & 26.8\\
\texttt{bge-gemma} & 53.0 & 54.0 & \bf 34.9 & 19.5 & 27.7 & 33.3 & 39.4 & 36.2 & 34.0 & 29.7\\
\midrule
\multicolumn{11}{c}{\it Fine-tuning (10k data on 1.5B/1.7B models)} \\
\midrule
Q2P/SFT & 46.6 & 51.2 & 30.7 & 19.7 & 29.4  & 38.0  & 39.7  & 35.6& 33.2& 31.1\\
\texttt{gte-Qwen}/ContrastiveFT & 57.2 & 61.4 & 29.1 & 18.6  & 26.4  & 32.9 & 39.2  & 35.5& 31.5& 29.0\\

\ours-\textsc{Format} & \bf 63.1 & \bf 66.1 & 22.6   & 17.6  & 23.4 & 30.0  & 32.2  & 30.1& 26.1& 25.9\\
\ours & 53.9 & 59.9 & 33.0 & \bf 23.1  & \bf 32.0  & \bf 41.2  & \bf 45.8  & \bf 43.0 & \bf 36.9& \bf 35.8\\
\bottomrule
\end{tabular}
\caption{Results on \textsc{ToolRet}, a benchmark of 35 datasets. Web$^\star$ denotes the zero-shot Web datasets.}
\label{tab:main}
\vspace{-10pt}

\end{table*}
\endgroup

\paragraph{Baselines.} 
\paragraph{Baselines.} We compare \ours against three categories of baselines. 
\textbf{Prompting methods} (using \texttt{Qwen3-30B-A3B-Instruct-2507}): (a) Q2E: direct query expansion~\citep{jagerman_query_2023}; (b) Q2D: query expansion by hypothetical tool generation~\citep{wang_query2doc_2023}; (c) HyDE: using averaged embeddings of generated tools~\citep{gao-etal-2023-precise}; (d) D2Q: document expansion~\citep{nogueira2019document}; and (e) Re-Invoke: Intent extraction then D2Q pipeline~\citep{chen-etal-2024-invoke}. 
\textbf{Re-ranking methods}: We re-score the top-20 results from \texttt{gte-Qwen} using cross-encoders, \texttt{bge-reranker-v2-m3} and \texttt{bge-reranker-v2-gemma}~\citep{chen-etal-2024-m3}. 
\textbf{Fine-tuning methods} (on the same data): (a) Q2P: A \texttt{Qwen3-1.7B} model fine-tuned to predict initial plans as expansions; and (b) ContrastiveFT: \texttt{gte-Qwen} fine-tuned via contrastive learning.

\paragraph{\ours.} We implemented two versions of \ours, both fine-tuned with the lightweight \texttt{Qwen3-1.7B}.
The proposed method is denoted as \ours, where we trained on 10k synthetic trajectories generated by \texttt{gpt-4.1-mini} \citep{achiam2023gpt} based on ToolRet's training set.
Additionally, we test whether additional formatting information aids in targeted domains, by additionally including the tool format definition as targets during SFT.
This variation is denoted as \ours-\textsc{Format}.
Detailed data generation procedure and training settings for both SFT and RLVR can be found in Appx~\ref{appx:training}.

\paragraph{Results.} 
Tab.~\ref{tab:main} presents the main retrieval results.
\ours and \ours-\textsc{Format} demonstrate superior performance across all settings. 
On in-domain splits, \ours-\textsc{Format} excels, improving over the base retriever by $\sim$20\% and outperforming ContrastiveFT by $\sim$6\%. 
This confirms that while standard contrastive fine-tuning improves representations, explicitly modeling tool schemas via generation greatly benefits task-specific applications. 
Crucially, for zero-shot generalization, \ours achieves the highest performance, surpassing all prompting, fine-tuning, and re-ranking baselines. 
Notably, \ours outperforms the state-of-the-art cross-encoder \texttt{bge-gemma} by $\sim$3\% for N@10 and $\sim$6\% for C@10. This indicates that interactive query planning is more effective at uncovering relevant tools than expensive post-hoc cross-attention re-scoring. 
Furthermore, \ours achieves these gains while being highly efficient: using a lightweight 1.7B model, it significantly outperforms query expansion methods that rely on prompting much larger (30B) models for multiple passes and avoids the high inference latency of cross-encoder re-ranking.

\subsection{Transfer to unseen base retriever}
\paragraph{Setup.} To validate the robustness of \ours, we evaluate the trained query planner \ours in a transfer setting where it interacts with different tool retriever models.
In this setting, \ours is trained with SFT data generated using the base retriever \texttt{gte-Qwen2-1.5B-instruct} and further fine-tuned via RLVR with the same retriever, but the query planner would be used at test time with a different retriever, i.e., changing the test environment.
This evaluates whether the planned queries capture universal semantic properties or are overfitted to the base retriever's specific embedding space.
We evaluate on a diverse set of general and tool-focused embedding models previously benchmarked on ToolRet.
We report the averaged gains in nDCG@10 and Completeness@10 over each corresponding base retriever.
Detailed breakdowns are provided in Appx.~\ref{appx:cross_eval}.

\paragraph{Results.}
As shown in Tab~\ref{tab:transfer}, \ours and \ours-\textsc{Format} consistently outperform baseline expansion methods, even when the inference environment differs from training. 
\ours-\textsc{Format} achieves 9$\sim$10\% gains in-domain, while \ours obtains 5$\sim$7\% improvements out-of-domain across diverse retrieval models.
This confirms that \ours's policy is highly robust and generalizable, providing significant value regardless of the underlying embedding model.

\begingroup
\setlength{\tabcolsep}{2.5pt}
\begin{table}[t!]
\small
\centering
\begin{tabular}{L{3cm} cc  cc}
\toprule
\multirow{2}{*}[-1ex]{\bf Method} 
    & \multicolumn{2}{c}{\bf In-Domain} 
    & \multicolumn{2}{c}{\bf 0-Shot} \\
    
\cmidrule(lr){2-3}
\cmidrule(lr){4-5}

& \scriptsize $\Delta$N@10 & \scriptsize $\Delta$C@10
& \scriptsize $\Delta$N@10 & \scriptsize $\Delta$C@10
\\
\midrule
Q2E & +1.6 & +1.9 & +2.2 & +2.8 \\
Q2P & +1.4 & +1.9 & +2.2 & +3.7 \\
D2Q & +3.3 & +3.3 & +2.8 & +2.8\\
Re-Invoke & +5.2 & +7.5 & +5.2 & +5.4 \\
\ours-\textsc{Format} & \bf +9.4 & \bf +9.6 & +1.6 & +4.2 \\
\ours & +5.6 & +6.8 & \bf +5.4 & \bf +7.1 \\
\bottomrule
\end{tabular}
\caption{Retriever transfer results on \textsc{ToolRet}. \ours is trained on \texttt{gte-qwen}-generated data, and used out-of-the-box directly with various retrievers at inference time. Average gains for NDCG@10 and Completeness@10 are reported.}
\label{tab:transfer}
\vspace{-10pt}

\end{table}
\endgroup
\subsection{End-to-end Tool Calling}
\paragraph{Benchmark and Setup.} To access the practical utility of \ours, we evaluate end-to-end tool-use performance of \texttt{Qwen3-30B} on two standard tool-use benchmarks, using \texttt{gte-Qwen} as the retriever. 
First, we use API-Bank~\citep{li2023api} (73 tools), testing on the retrieval-focused Level-2 subset and the Level-1 subset configured to force tool search.
We report the average accuracy based on the top retrieved tool.

Next, we evaluated on StableToolBench~\citep{guo-etal-2025-stabletoolbench} (STB), the stable version of the most widely-used ToolBench~\citep{qin2024toolllm} in the past few years.
We evaluated on the subsets I2-Category (13k tools) and I3-Instruction (1.6k tools).
Compared to API-Bank, these tasks are much more complex and compositional, and requiring multiple steps of reasoning and tool call execution.
We follow the official ReAct~\citep{yao2023react} inference pipeline, and report the solvable pass rate.

\begingroup
\setlength{\tabcolsep}{2.5pt}
\begin{table}[t!]
\small
\centering
\begin{tabular}{L{2.6cm} cc  cc}
\toprule
\multirow{3}{*}[-1.3ex]{\bf Retrieval Method} 
    & \multicolumn{2}{c}{\bf API-Bank} 
    & \multicolumn{2}{c}{\bf STB} \\
    
\cmidrule(lr){2-3}
\cmidrule(lr){4-5}

& \multicolumn{2}{c}{\tt Qwen3-30B} & \multicolumn{2}{c}{\tt Qwen3-30B-ReAct} \\
\cmidrule(lr){2-3}
\cmidrule(lr){4-5}
& Level-1 & Level-2 & I2-Cat & I3-Inst\\
\midrule
Base Retriever & 30.2 & 57.8 & 51.9  & 40.3  \\
Q2E & 31.3 & 59.5  & 48.3 & 38.6  \\
Q2D & 31.8 & 58.5  & 51.6 & 40.4  \\
Re-Invoke & 30.5 & 60.5  & 57.9 & 28.6 \\
Q2P & 30.1 & 56.1 &51.9  & 41.0  \\
\ours & \bf 33.0  & \bf 60.7  & \bf 60.2 & \bf 48.3  \\
\bottomrule
\end{tabular}
\caption{End-to-end tool-calling performance for \texttt{Qwen3-30B} with different retrieval methods on API-Bank and StableToolBench (STB). Accuracy is reported for API-Bank and is Solvable Pass Rate reported for StableToolBench.}
\vspace{-10pt}
\label{tab:e2e}

\end{table}
\endgroup
\paragraph{Results.}
Tab.~\ref{tab:e2e} highlights the downstream impact of improved retrieval.
On the smaller API-Bank, \ours achieves the highest accuracy with a $\sim$3\% accuracy gain.
This is notable as \ours is designed to operate and excel for large toolsets and not for one as small as 73 tools, which suggests utility even in simpler scenarios.
On the larger and more complex STB, \ours achieves substantial gains of 8$\sim$9\% tool-use pass rate over the base retriever.
Notably, \ours is the most consistent method; unlike baselines that occasionally degrade performance on some subsets, our planner provides robust context that reliably aids the downstream agent.
This is achieved with a lightweight 1.7B planner, and importantly, generalizes well even though STB utilizes a tool embedding format different from which \ours was trained on, showcasing its robustness again.

\subsection{Ablation Studies}
We conduct the following ablation studies to validate our design choices, with results summarized in Tab.~\ref{tab:ablation}.

\paragraph{Prompting vs SFT vs RLVR.}
We first investigate whether explicit training is necessary by comparing \ours against \ours-\textsc{Prompting}, a version of our pipeline that uses \texttt{Qwen3-30B} to plan queries zero-shot, and further compare against only bootstrapping with synthesized data with SFT.
We find that \ours-\textsc{Prompting} already achieves big improvements, especially on zero-shot generalization, where it outperforms all prompting baselines shown earlier.
This demonstrates the effectiveness and necessity of the query planning framework.
However, by fine-tuning with the proposed data generation method, we can further improve performance, especially within the domain.
On the other hand, we observe that the bulk of in-domain performance gains come from the SFT stage, while RLVR is critical for refining the policy to achieve optimal out-of-domain performance.

\begingroup
\setlength{\tabcolsep}{2.5pt}
\begin{table}[t!]
\small
\centering
\begin{tabular}{L{3.5cm} cc  cc}
\toprule
\multirow{2}{*}[-1ex]{\bf Method} 
    & \multicolumn{2}{c}{\bf In-Domain} 
    & \multicolumn{2}{c}{\bf 0-Shot} \\
    
\cmidrule(lr){2-3}
\cmidrule(lr){4-5}

& N@10 & C@10
& N@10 & C@10
\\
\midrule
Base Retriever (\texttt{gte-Qwen}) & 43.3 & 47.8 &  29.8 & 27.0 \\
\midrule
\multicolumn{5}{c}{\it Prompting vs SFT vs RLVR} \\
\midrule
\ours-\textsc{Prompting} & 45.6 & 52.4 & 34.0& 33.0\\
\ours-\textsc{SFT} & 52.4 & 57.9 & 35.4 & 34.4 \\
\ours-\textsc{RLVR} & \bf 53.9 & \bf 59.9 & \bf 36.9& \bf 35.8\\
\midrule
\multicolumn{5}{c}{\it Aggregation Methods} \\
\midrule
Reciprocal Rank Fusion & 50.9  & 56.3 & 35.4 & 34.5 \\
Multi-view Fusion & 54.1 & 60.2 & 37.0 & 35.5 \\
Peak-rank & 53.9 & 59.9 & 36.9& \bf 35.8\\
Cross-encoder (\texttt{bge-gemma}) & \bf 58.2 & \bf 62.2 & \bf 37.3& \bf 35.8\\
\midrule
\multicolumn{5}{c}{\it Retrieval with/without user query} \\
\midrule
\ours w/o user query & 52.3 & 57.7 & 34.8& 33.3\\
\ours & \bf 53.9 & \bf 59.9 & \bf 36.9& \bf 35.8\\
\bottomrule
\end{tabular}
\caption{Ablation studies.}
\vspace{-12pt}
\label{tab:ablation}
\end{table}
\endgroup
\paragraph{Choice of Aggregation Method.}
Next, we compare our chosen aggregation method, the simple peak-rank fusion, to the widely-used reciprocal rank fusion (RRF)~\citep{10.1145/1571941.1572114}, and a multi-view fusion method that joins retrieved lists from extracted user intents~\citep{chen-etal-2024-invoke}.
Peak-rank outperforms RRF by a wide margin, likely by mitigating the frequency bias for sub-tasks that require more query attempts, and performs on par with the more complicated multi-view fusion.
We also tested using the \texttt{bge-reranker-v2-gemma} cross-encoder for fusing the lists, useful when more compute is available, which further boosts in-domain performance.

\paragraph{Importance of user query.}
Finally, we test whether we can rely solely on the queries for sub-tasks.
We find removing that information degrades the performance, since decomposition may be incomplete or erroneous, and including the original user query could act as an anchor.
We also note that all query expansion baselines we implemented do include the original query to achieve their best (and reported) results.

\subsection{Qualitative Analysis}
Figure~\ref{fig:qual} visualizes \ours discovering implicit dependencies.
The task ``modify password'' requires a token (via \texttt{GetUserToken}), a prerequisite absent from the user query.
While baselines fail by focusing solely on keywords, \ours identifies this relation and successfully retrieves the authentication tool alongside the target.
This highlights \ours's capacity to uncover latent inter-tool dependencies.
Four additional examples taken from the test set are analyzed in Appx.~\ref{appx:qualitative}.
\begingroup
\setlength{\tabcolsep}{2.5pt}
\begin{figure}[t!]
\fontsize{7}{9}\selectfont
\renewcommand{\arraystretch}{1.15}
\centering
\resizebox{\linewidth}{!}{
\begin{tabular}{lL{3.2cm}L{2.5cm}}
\toprule
\rowcolor{gray!8}
\it User Query&\multicolumn{2}{l}{Modify password} \\
\multirow{2}{*}[-0.6ex]{\it Targets} &\multicolumn{2}{l}{\tt ModifyPassword(token,old\_password,new\_password)} \\
 &\multicolumn{2}{l}{\tt GetUserToken(username,password)} \\
\midrule
\bf Method & \bf Query & \bf Top-5 Retrieved \\
\midrule
\rowcolor{gray!8}
\ours & \texttt{<sub\_goal> Modify a user's password by processing the username and new password inputs </sub\_goal>} tool for modifying user password with inputs username string and new password string, password modification, \colorbox{yellow!50}{user authentication}, password update & \tt ModifyPassword, ForgotPassword, RegisterUser, \colorbox{green!30}{GetUserToken}, OpenBankAccount\\
dense & modify password & \tt ModifyPassword, ForgotPassword, ModifyAgenda, ModifyRegistration, OpenBankAccount\\
\rowcolor{gray!8}
Q2E & tools to modify password, change password, update password, reset password, password management tools, password reset utilities, password change methods, secure password modification tools & \tt ModifyPassword, ForgotPassword, ModifyAgenda, ModifyAlarm, OpenBankAccount \\
Q2D& \tt \{"name": "modify\_password", "description": "A tool to update or change a user's password.", "parameters": \{"new\_password": \{"description": "The new password to set for the user.", "type": "string"\}, ... & \tt ModifyPassword, ForgotPassword, ModifyAgenda, ModifyAlarm, RegisterUser\\
\rowcolor{gray!8}
Q2P& Given a `password modification` task, retrieve tools that allow users to change their passwords by processing the username and new password inputs & \tt ModifyPassword, ForgotPassword, ModifyAgenda, ModifyRegistration, OpenBankAccount \\
Re-Invoke& modify password & \tt ModifyPassword, ForgotPassword, ModifyAlarm, ModifyAgenda, ModifyReminder \\
\bottomrule
\end{tabular}
}
\caption{\ours discovering dependencies.}
\vspace{-16pt}
\label{fig:qual}
\end{figure}
\endgroup

\section{Conclusion}
We introduce \ours, a framework that formulates tool retrieval as an iterative planning process. 
By decomposing instructions and leveraging interactive feedback, \ours bridges the semantic gap between abstract user intents and technical tool documentation. 
This is enabled through a robust training pipeline using synthetic trajectories and RLVR.
Empirically, \ours outperforms state-of-the-art baselines on ToolRet, demonstrating superior zero-shot generalization. 
Furthermore, our analysis confirms that the learned planning policy is highly robust, transferring effectively to unseen retrieval models and significantly improving success rates in downstream end-to-end agentic tasks.
Lightweight and modular, \ours offers a scalable solution for robust agents operating over massive tool libraries.

\pagebreak
\section*{Acknowledgments}
This research is supported by the Centre for Perceptual and Interactive Intelligence (CPII) Ltd. under the Innovation and Technology Commission’s InnoHK Scheme.

\section*{Limitations}
Our synthetic data generation utilized a single base retriever due to compute constraints, implicitly biasing the planner to a specific embedding space; future work should explore multi-retriever synthesis to foster a more robust, agnostic policy.
Additionally, \ours resolves inter-tool dependencies implicitly via interaction, whereas integrating explicit Tool Knowledge Graphs could guide the planning process more efficiently than trial-and-error. 
In terms of scope, our framework targets massive, dynamic tool libraries, meaning the complexity of iterative planning may be unnecessary for smaller toolsets compared to direct prompting. 
Finally, developing an adaptive planning mechanism that dynamically bypasses or simplifies decomposition for simple queries would further optimize the latency-performance trade-off.

\section*{Ethics Statement}
We used publicly available models and datasets for training and evaluation, and did not collect data or any personal information in this work. 
Self-generated data by these public models were used to train a generation model, which has a chance of producing harmful or hallucinated content, even when generation is conditioned on curated public data inputs and filtered with ground truth targets from public data. 
They could potentially be misused and pose ethical risks typical of large language models when deployed in real-world applications, if not thoroughly audited.

\bibliography{anthology1,anthology2,custom,tool_retrieval,retrieval,rl}

\appendix

\section{Dataset Details and Licenses}
\label{appx:data}
We use ToolRet for tool retrieval evaluation, which aggregates 35 previously published datasets; full dataset statistics are documented in~\citet{shi-etal-2025-retrieval}. ToolRet is licensed under Apache License 2.0.
For end-to-end tool use evaluation, we used API-Bank~\citep{li2023api} and StableToolBench~\citep{guo-etal-2025-stabletoolbench}, which are licensed under MIT License and Apache License 2.0, respectively.

\section{Baseline Implementation Details and Prompts}
\label{appx:baseline}
We include the implementation details for baselines used in our evaluation in Tab.~\ref{tab:baseline_details}.
We include all prompts at the end of the appendices.

\begingroup
\setlength{\tabcolsep}{2.5pt}
\begin{table*}[t!]
\small
\centering
\begin{tabular}{L{3.7cm}llL{4.5cm}l}
\toprule
\bf Baseline & \bf Method & \bf Base Model & \bf Details & \bf Prompts \\
\midrule
Q2E~\citep{jagerman_query_2023} & ZS/PRF Prompting & \texttt{Qwen3-30B}& Dense fusion of $N=5$ query expansions concatenated to user query. & See Figs.~\ref{fig:prompt_q2e},~\ref{fig:prompt_q2e_prf} \\
\midrule
Q2D~\citep{wang_query2doc_2023} & ZS/PRF Prompting &\texttt{Qwen3-30B} & Dense fusion of $N=5$ query expansions (hypothetical documents) concatenated to user query. & See Figs.~\ref{fig:prompt_q2d},~\ref{fig:prompt_q2d_prf} \\
\midrule
HyDE~\citep{gao-etal-2023-precise} & ZS/PRF Prompting & \texttt{Qwen3-30B}& Dense fusion of $N=5$ hypothetical documents \emph{and} user query. & See Figs.~\ref{fig:prompt_q2d},~\ref{fig:prompt_q2d_prf} \\
\midrule
D2Q~\citep{nogueira2019document} & ZS Prompting &\texttt{Qwen3-30B} & Averaged embeddings of $N=10$ hypothetical queries concatenated to tool documentation. & See Fig.~\ref{fig:prompt_d2q} \\
\midrule
Re-Invoke~\citep{chen-etal-2024-invoke} & FS Prompting & \texttt{Qwen3-30B}& Same embeddings as D2Q. & See Figs.~\ref{fig:prompt_d2q},~\ref{fig:prompt_reinvoke} \\
\midrule
Q2P & Supervised Fine-tuning & \texttt{Qwen3-1.7B}& 10k ToolRet-train data; trained with \href{https://huggingface.co/docs/trl/index}{Huggingface TRL}'s \texttt{SFTTrainer} with the following configurations: 3 epochs, warmup ratio $0.03$, batch size $64$, learning rate $2e-5$ & N/A \\
\midrule
\texttt{gte-Qwen}/ContrastiveFT & Noise-contrastive Fine-tuning & \texttt{gte-Qwen}& 10k ToolRet-train data; trained with \href{https://github.com/FlagOpen/FlagEmbedding}{FlagEmbedding} with the following configuration: \texttt{decoder\_only, train\_group\_size=8, learning\_rate=1e-6, num\_train\_epochs=2, warmup\_ratio=0.1}. & N/A \\
\midrule
\ours-\textsc{Prompting} & ZS Prompting &\texttt{Qwen3-30B}& Sampling temperature set to $0$ for planning and $0.5$ for query generation. & See Fig.~\ref{fig:prompt_toolqpp} \\
\bottomrule
\end{tabular}
\caption{Baseline details.}
\label{tab:baseline_details}

\end{table*}
\endgroup

\section{\ours Training Details}
\label{appx:training}
\subsection{Data}
For training, we subsampled the training data provided in~\citet{shi-etal-2025-retrieval}, which is compiled from 3 datasets: ToolBench~\citep{qin2024toolllm}, ToolACE~\citep{liu2025toolace}, and APIGen~\cite{liu2024apigen}. 
The original training set contains roughly 200k instances, which we subsample and select 10k for data generation and training.
\subsection{Query Sequence Generation}
Detailed illustration of the algorithm used to generate query trajectories can be found in Alg.~\ref{alg:datagen}.
Probability $p$ of keeping failed attempts is set to $0.4$, and the rank threshold $r_\tau$ is set to 5.
\texttt{gpt-4.1-mini} is used as the teacher model with default sampling parameters.

In Fig.~\ref{fig:training_example}, we show an example query trajectory sequence, which is used for SFT with maximum likelihood with teacher forcing.
Prompts for the teacher model used in the described algorithm can be found in Fig.~\ref{fig:prompt_toolqp} at the end of appendices.
\subsection{SFT}
We train \ours and \ours-\textsc{Format} with \href{https://huggingface.co/docs/trl/index}{Huggingface TRL}'s \texttt{SFTTrainer}, with the following configurations:
3 epochs, learning rate of 2e-5, 150 warmup steps with cosine schedule, batch size of 32, weight decay of 0.01, and a maximum sequence length of 16384.
We include the top $k=5$ retrieved tools as retrieval feedback for teacher forcing.

\subsection{RLVR}
We used the \href{https://github.com/PrimeIntellect-ai/verifiers}{\tt verifiers} library to implement GRPO training~\citep{brown_verifiers_2025} since it supported multi-step rollout and training.
We use a setup of: 500 training steps, learning rate of 1e-6, (effective) batch size of 256, early stopping with dev set, warmup steps of 143, kl regularization of 0.02, group size of 4, $\mu$ set to 2, top $k=5$ feedback per turn with 10 maximum query turns.

For reward shaping, we define $\mathcal{R}_\text{retrieval}=\beta_{1,n}\Delta N+\beta_{1,r}\Delta R$, where $\Delta N$ is the nDCG@5 difference between the final aggregated retrieval from the sequence and the baseline retrieval with $\mathrm{Concat}(q, \mathcal{P})$, and similarly defined for recall difference $\Delta R$.
The format reward $\mathcal{R}_\text{format}=\beta_{2,f}R_f+\beta_{2,s}R_s$ consists of $R_f$, the fraction of responses in correct formatting, and $R_s$, whether rollout sequence ends in the \texttt{<stop\_retrieval>} (end of query generation) token.
Finally, $\mathcal{R}_\text{plan}=\mathrm{Sim}(P, \mathcal{P})$ is the cosine similarity score between the first turn plan prediction $P$ and reference plan $\mathcal{P}$ using an embedding model. 
We use \texttt{gte-Qwen} for convenience.
The reward weights are set to: $\beta_{1,n}=5.0,\beta_{1,r}=2.5,\beta_{2,f}=1.5,\beta_{2,s}=0.6,\beta_3=1.0$.

\section{Details for Retriever Transfer Evaluation}
\label{appx:cross_eval}
For retriever transfer evaluation, we test on 5 widely-used embedding models previously tested on ToolRet: \texttt{bge-large-en-v1.5}, \texttt{e5-base-v2}, \texttt{e5-mistral-7b-instruct}, \texttt{ToolRet-bge-large-en-v1.5} and \texttt{ToolRet-e5-base-v2}, with the last two being fine-tuned versions of the first two models using the full 200k training data in ~\citet{shi-etal-2025-retrieval} by contrastive training.
Details for each model are provided in Tab.~\ref{tab:retriever_details}.
In the main text, we reported the averaged improvements of all 5 models for each baseline method and \ours.
Additionally, we provide in Tab.~\ref{tab:ret_transfer} detailed results for each model-method pairing for each category in ToolRet.
\begingroup
\setlength{\tabcolsep}{5.5pt}
\begin{table}[t!]
\small
\centering
\begin{tabular}{lc}
\toprule
\multirow{2}{*}[-1ex]{\bf Retrieval Method} 
    & \bf API-Bank \\
\cmidrule(lr){2-2}
& Level-1 \\
\midrule
Base Retriever & 47.6 \\
Q2E & 48.3 \\
Q2D & 46.6 \\
Re-Invoke & 41.7 \\
Q2P & 48.0\\
\ours & \bf 53.2 \\
\bottomrule
\end{tabular}
\caption{End-to-end tool-calling performance for \texttt{Qwen3-30B} on API-Bank Level-1 for the alternative setting where the initial context for the downstream LLM contains retrieved APIs.}
\vspace{-10pt}
\label{tab:apibank_lvl1}
\end{table}
\endgroup

\begingroup
\setlength{\tabcolsep}{2.5pt}
\begin{table*}[t!]
\small
\centering
\begin{tabular}{llll}
\toprule
\bf Model & \bf Reference & \bf \#Params & \bf Source \\
\midrule
\texttt{gte-Qwen2-1.5B-Instruct}&~\citet{li2023towards} & 1.5B & \href{https://huggingface.co/Alibaba-NLP/gte-Qwen2-1.5B-instruct}{Alibaba-NLP/gte-Qwen2-1.5B-instruct} \\
\texttt{bge-large-en-v1.5}&~\citet{bge_embedding} & 335M & \href{https://huggingface.co/BAAI/bge-large-en-v1.5}{BAAI/bge-large-en-v1.5} \\
\texttt{ToolRet-bge-large-en-v1.5}&~\citet{shi-etal-2025-retrieval} & 335M & \href{https://huggingface.co/mangopy/ToolRet-trained-bge-large-en-v1.5}{mangopy/ToolRet-trained-bge-large-en-v1.5} \\
\texttt{e5-base-v2}&~\citet{wang2022text} & 110M & \href{https://huggingface.co/intfloat/e5-base-v2}{intfloat/e5-base-v2} \\
\texttt{ToolRet-e5-base-v2}&~\citet{shi-etal-2025-retrieval} & 110M & \href{https://huggingface.co/mangopy/ToolRet-trained-e5-base-v2}{mangopy/ToolRet-trained-e5-base-v2} \\
\texttt{e5-mistral-7b-instruct}&~\citet{wang-etal-2024-improving-text} & 7B & \href{https://huggingface.co/intfloat/e5-mistral-7b-instruct}{intfloat/e5-mistral-7b-instruct} \\
\bottomrule
\end{tabular}
\caption{Base retriever details.}
\label{tab:retriever_details}

\end{table*}
\endgroup

\section{Details for End-to-end Tool Use Evaluation}
\label{appx:e2e_eval}
We use the official evaluation pipelines for both API-Bank and StableToolBench.
For both datasets, we report results with the average of 3 runs.
\texttt{gte-Qwen} is used as the retriever, while all baselines are the same as previous settings.
The fine-tuned models, including \ours, are tested directly on the datasets without further adaptation.

For API-Bank, level-2 is used as-is.
For level-1, the standard evaluation is unrealistic since it gives the list of all APIs used for a dialogue at the beginning of dialogue, making it much easier for the LLM to choose from.
Under the setting used in the main text, we modify it such that every turn where the target is an API call, we replace the initial list such that the LLM can only observe \texttt{ToolSearcher}, the tool retriever, and the original API must be retrieved to be deemed correct.
Alternatively, in Tab.~\ref{tab:apibank_lvl1}, we also show the setting where the initial semi-ground-truth list is replaced with a retrieved top-$k$ list using the initial user query, which is quite common in tool retrieval pipelines, including the STB evaluation below.
The tool retrieval step is forced as a required step before the LLM acts in this case.
We set $k=5$ for this set of experiments.
Compared to the other setting shown in the main text, we observe a even more significant outperformance by \ours over baselines consistently, with an improvement of $\sim$5\% over all baselines.

For StableToolBench, we use \texttt{Qwen-30B} as the ToolEval LLM judge, and the \texttt{stabletoolbench/MirrorAPI-Cache} model as the tool simulation model.
We follow the official inference script default and set $k=5$ to including the 5 top-retrieved tools.

\section{Additional Qualitative Studies}
\label{appx:qualitative}
In the main text, we showed an example of \ours successfully retrieving all relevant tools and discovering implicit tool dependencies compared to other methods.
Here, we show four additional examples taken from ToolRet's test set.
Figs.~\ref{fig:qual2} and \ref{fig:qual3} show examples of \ours correctly finding relevant tools by planning a single sub-goal and generating one query.
For the former, \ours managed to focus on general books and include crucial keywords such as ``publication year'' and ``genre'' that effectively guided the retriever, while other methods mostly retrieved Bible-related tools due to the frequent use of the term ``book'' in Bible-related tool names and documentations.
For the latter, the expansion baselines largely fail to retrieve the correct targets as they focus on the high-level ``calculate number of trees planted'' and retrieved quite a few plant-related or calculation tools; \ours, on the other hand, accurately finds the target tool by articulating the plan correctly and generates queries that target such a calculation tool specifically for organizations.
In Figs.~\ref{fig:qual4} and \ref{fig:qual5}, we analyze and compare \ours to different methods on complex queries that require more fine-grained task decomposition.
In the former case, \ours decomposes the user query into two distinct sub-goals, and retrieves all three targets perfectly; most other expansion methods fail, in comparison, and focuses on tools related to the topic ``exercise'' or generic video-related tools, due to complexity of the user query and the usage of only one single combined query for representation; and while Re-Invoke decomposes the task into three queries, it fails to find relevant tools for two out of the three query attempts.
For the final example, all other expansion baselines other than Re-Invoke retrieved only RL-related models endpoints rather than the research paper and repositories the user requested; Re-Invoke manages to find paper-related tools (although not labeled as targets), but fails to rank repository-searching tools high.
In comparison, \ours's sub-goals and queries are much more detailed and targeted, and results in accurate retrieval. 
We also note that \ours spends an extra step for the second sub-goal to confirm, likely due to uncertainty in the feedback, using a slight modification of the query in the previous turn; in this case, our aggregation algorithm does not bias towards tools that appear in more turns, as designed, so that the top-5 are not populated with just GitHub-related tools.

\begingroup
\setlength{\tabcolsep}{2.5pt}
\begin{figure*}[t!]
\footnotesize
\renewcommand{\arraystretch}{1.15}
\centering
\begin{tabular}{lP{6.4cm}P{8.1cm}}
\toprule
\rowcolor{gray!8}
\it User Query&\multicolumn{2}{l}{Can I have a summary of a thriller book by Dan Brown that was published in 2003?} \\
\it Targets &\multicolumn{2}{l}{\tt BookFinder(genre, author, year), BookSummary(book\_name)} \\
\midrule
\bf Method & \bf Query & \bf Top-5 Retrieved \\
\midrule
\rowcolor{gray!8}
\ours & \begin{tabular}[t]{@{}p{\linewidth}@{}}
\texttt{<sub\_goal> Retrieve a summary of a thriller book by a specified author and publication year. </sub\_goal>}\\ 
tool for retrieving book summary by author and \colorbox{yellow!50}{publication year}, parameters: author name, publication year, category: book summary, \colorbox{yellow!50}{genre}: thriller 
\end{tabular}& {\tt\begin{tabular}[t]{@{}p{\linewidth}@{}}
\colorbox{green!30}{BookSummary(book\_name)}\\
TopBooks(author\_info, num\_of\_books)\\
Bible-Memory-Verse-Flashcard-Search-Term-Verse-\\Address-Summary(term1)\\
\colorbox{green!30}{BookFinder(genre, author, year)}\\
BookTitle(author, genre)\\
\end{tabular}} \\
dense & Can I have a summary of a thriller book by Dan Brown that was published in 2003? & {\tt\begin{tabular}[t]{@{}p{\linewidth}@{}} BibleSearch.GetChapterbyBookName(bookName, chapterId)\\
GetChapter(Book, chapter)\\
UncoveredTreasureVerse.verse(verse)\\
GetBookDetails(bookName)\\
CompleteStudyBible.FullChapterAPI(chapter, translation, book)\end{tabular}}\\
\rowcolor{gray!8}
Q2E & summary of a thriller book by Dan Brown published in 2003 & {\tt\begin{tabular}[t]{@{}p{\linewidth}@{}}
GetBookDetails(bookName)\\
BibleSearch.GetChapterbyBookName(bookName, chapterId)\\
UncoveredTreasureVerse.verse(verse)\\
HolyBible.GetChapter(Book, chapter)\\
BibleSearch.GetBookByName(bookName)
\end{tabular}} \\
Q2D & \tt \{"name": "get\_book\_summary", "description": "Retrieves a summary of a thriller book by Dan Brown published in 2003.", "parameters": \{"book\_title": \{"description": "The title of the book by Dan Brown published in 2003.", "type": "string"\}\}\} & {\tt\begin{tabular}[t]{@{}p{\linewidth}@{}}
\colorbox{green!30}{BookSummary(book\_name)}\\
GetBookDetails(bookName)\\
BibleSearch.GetBookByName(bookName)\\
BookTitle(author, genre)\\
GetBookInformation(book\_Id)
\end{tabular}} \\
\rowcolor{gray!8}
Q2P & Given a `book summary` task, retrieve tools that extract summaries based on book title and publication year by processing these inputs to provide the requested information. & {\tt\begin{tabular}[t]{@{}p{\linewidth}@{}}
Lead-3(text)\\
\colorbox{green!30}{BookSummary(book\_name)}\\
GetBookDetails(bookName)\\
find\_page\_number()\\
SearchTerm.ChapterAddressSummary(first\_book, second\_book, term1)
\end{tabular}} \\
Re-Invoke & summarize a thriller book by Dan Brown published in 2003 & {\tt\begin{tabular}[t]{@{}p{\linewidth}@{}}
BookInspector()\\
find\_page\_number()\\
generateThrillerPlot(protagonist, antagonist)\\
IdentifyBook(title)\\
FindBook(title)
\end{tabular}} \\
\bottomrule
\end{tabular}
\caption{Retrieval comparison for example taken from ToolRet's test set.}
\label{fig:qual2}
\end{figure*}
\endgroup
\begingroup
\setlength{\tabcolsep}{2.5pt}
\begin{figure*}[t!]
\footnotesize
\renewcommand{\arraystretch}{1.15}
\centering
\begin{tabular}{lP{6.4cm}P{8.1cm}}
\toprule
\rowcolor{gray!8}
\it User Query&\multicolumn{2}{P{14.5cm}}{An environmental agency examined how many trees were planted by different organizations. In all, how many trees were planted by Let it Grow and Heal the Earth?} \\
\it Targets &\multicolumn{2}{>{\raggedright\arraybackslash}p{14.5cm}}{\tt calculate\_total\_items\_for\_organizations(df, organization\_col, item\_count\_col, organization\_list)} \\
\midrule
\bf Method & \bf Query & \bf Top-5 Retrieved \\
\midrule
\rowcolor{gray!8}
\ours & \begin{tabular}[t]{@{}p{\linewidth}@{}}
\tt <sub\_goal> Calculate the total number of trees planted by specific organizations by processing the organization names as inputs. </sub\_goal>\\
tool to \colorbox{yellow!50}{calculate total number} of trees planted by specific organizations, input parameters: \colorbox{yellow!50}{organization names}, output: total tree count, category: environmental data, tree planting statistics
\end{tabular}& {\tt\begin{tabular}[t]{@{}p{\linewidth}@{}}
\colorbox{green!30}{\parbox[t]{\linewidth}{\raggedright\arraybackslash calculate\_total\_items\_for\_organizations(df, \newline organization\_col, item\_count\_col, organization\_list)}}\\
calculate\_num\_trees(yard\_length, distance\_between\_trees)\\
count\_trees(leaves, threshold)\\
count\_gardens(plants)\\
count\_leaves\_1\_to\_7(plot)
\end{tabular}} \\
dense & An environmental agency examined how many trees were planted by different organizations. In all, how many trees were planted by Let it Grow and Heal the Earth? & {\tt\begin{tabular}[t]{@{}p{\linewidth}@{}}
IndoorPlants()\\
PlantRecommender()\\
calculate\_num\_trees(yard\_length, distance\_between\_trees)\\
GetAllPlants()\\
count\_gardens(plants)
\end{tabular}} \\
\rowcolor{gray!8}
Q2E & Total number of trees planted by Let it Grow and Heal the Earth combined & {\tt\begin{tabular}[t]{@{}p{\linewidth}@{}}
calculate\_num\_trees(yard\_length, distance\_between\_trees)\\
IndoorPlants()\\
CarbonFootprint\_TreeEquivalent(weight, unit)\\
PlantRecommender()\\
count\_gardens(plants)
\end{tabular}} \\
Q2D & \tt \{"name": "sum\_trees\_planted", "description": "Calculates the total number of trees planted by two specified organizations.", "parameters": \{"organization1": \{"description": "The name of the first organization.", "type": "string"\}, "organization2": \{"description": "The name of the second organization.", "type": "string"\}\}\} & {\tt\begin{tabular}[t]{@{}p{\linewidth}@{}}
calculate\_num\_trees(yard\_length, distance\_between\_trees)\\
calculate\_sum(num1, num2)\\
CarbonFootprint\_TreeEquivalent(weight, unit)\\
calculateSum(num1, num2)\\
treeequivalent(weight, unit)
\end{tabular}} \\
\rowcolor{gray!8}
Q2P & Given a `tree planting query`, retrieve tools that can provide the number of trees planted by specific organizations by processing the organization name as input. & {\tt\begin{tabular}[t]{@{}p{\linewidth}@{}}
CharityTool()\\
PlantRecommender()\\
GetAllPlants()\\
\colorbox{green!30}{\parbox[t]{\linewidth}{\raggedright\arraybackslash calculate\_total\_items\_for\_organizations(df, \newline organization\_col, item\_count\_col, organization\_list)}}\\
calculate\_num\_trees(yard\_length, distance\_between\_trees)
\end{tabular}} \\
Re-Invoke & \begin{tabular}[t]{@{}p{\linewidth}@{}}
determine the total number of trees planted by Let it Grow\\\\
determine the total number of trees planted by Heal the Earth\\\\
calculate the combined total number of trees planted by Let it Grow and Heal the Earth
\end{tabular}  & {\tt\begin{tabular}[t]{@{}>{\raggedright\arraybackslash}p{\linewidth}@{}}
calculate\_num\_trees(yard\_length, distance\_between\_trees)\\
CarbonFootprint\_TreeEquivalent(weight, unit)\\
count\_gardens(plants)\\
calculate\_total\_germination\_rate(seeds\_1, seeds\_2, germination\_rate\_1, germination\_rate\_2)\\
case1\_count()
\end{tabular}} \\
\bottomrule
\end{tabular}
\caption{Retrieval comparison for example taken from ToolRet's test set.}
\label{fig:qual3}
\end{figure*}
\endgroup

\begingroup
\setlength{\tabcolsep}{2.5pt}
\begin{figure*}[t!]
\footnotesize
\renewcommand{\arraystretch}{1.15}
\centering
\begin{tabular}{lP{6.4cm}P{8.1cm}}
\toprule
\rowcolor{gray!8}
\it User Query&\multicolumn{2}{P{14.5cm}}{I have an article about the benefits of daily exercise. I want to rewrite the article to make it unique, expand the content to include more detailed descriptions, and find a related video. Here's the original article text: 'Regular exercise is essential for maintaining physical and mental health. It can improve mood, increase energy levels, and help prevent various diseases.'} \\
\it Targets &\multicolumn{2}{l}{\tt ArticleSpinner(text), TextExpander(text), VideoSearch(text)} \\
\midrule
\bf Method & \bf Query & \bf Top-5 Retrieved \\
\midrule
\rowcolor{gray!8}
\ours & \begin{tabular}[t]{@{}p{\linewidth}@{}}
\tt <sub\_goal> Rewrite and expand article content based on the original article text</sub\_goal>\\
tool for rewriting and expanding article content, input: article text string, output: rewritten and expanded unique article text, content rewriting, article expansion, text processing\\
\tt <sub\_goal> Find related videos based on the article content. </sub\_goal>\\
tool for finding related videos based on article content, input: article text string, output: related video list, video retrieval, content-based video search, article content video linking
\end{tabular} & {\tt\begin{tabular}[t]{@{}p{\linewidth}@{}}
\colorbox{green!30}{ArticleSpinner(text)}\\
VideoSummarizeTool(link)\\
\colorbox{green!30}{VideoSearch(text)}\\
\colorbox{green!30}{TextExpander(text)}\\
TextParaphraser(text)\\
\end{tabular}} \\
dense & I have an article about the benefits of daily exercise. I want to rewrite the article to make it unique, expand the content to include more detailed descriptions, and find a related video. Here's the original article text: 'Regular exercise is essential for maintaining physical and mental health. It can improve mood, increase energy levels, and help prevent various diseases.' & {\tt\begin{tabular}[t]{@{}p{\linewidth}@{}}
exercise.video\_library(category, difficulty\_level)\\
Planfit()\\
getAllExercises()\\
getExerciseDetails(id)\\
v1\_exercises(offset, muscle, type, difficulty, name)
\end{tabular}} \\
\rowcolor{gray!8}
Q2E & Rewrite and expand the article on the benefits of daily exercise with detailed descriptions, and find a relevant educational video on the topic. & {\tt\begin{tabular}[t]{@{}p{\linewidth}@{}}
exercise.video\_library(category, difficulty\_level)\\
Planfit()\\
VideoSummarizeTool(link)\\
video\_highlight(link)\\
getAllExercises()
\end{tabular}} \\
Q2D & \tt \{"name": "rewrite\_and\_enhance\_content", "description": "Rewrites the provided article to make it unique, expands the content with detailed descriptions, and includes a related video recommendation.", "parameters": \{"original\_text": \{"description": "The original article text about the benefits of daily exercise.", "type": "string"\}\}\} & {\tt\begin{tabular}[t]{@{}p{\linewidth}@{}}
exercise.video\_library(category, difficulty\_level)\\
Planfit()\\
video\_highlight(link)\\
VideoSpeedChanger(video, text)\\
getExerciseDetails(id)
\end{tabular}} \\
\rowcolor{gray!8}
Q2P & Given a `content rewriting and video search` task, retrieve tools that can rewrite and expand text content while also finding related videos based on the given topic and context. The tools should process the original article text and return the rewritten content and a list of related videos. & {\tt\begin{tabular}[t]{@{}p{\linewidth}@{}}
exercise.video\_library(category, difficulty\_level)\\
\colorbox{green!30}{ArticleSpinner(text)}\\
video\_highlight(link)\\
Planfit()\\
VideoSummarizeTool(link)
\end{tabular}} \\
Re-Invoke & \begin{tabular}[t]{@{}p{\linewidth}@{}}
rewrite the article about the benefits of daily exercise to make it unique\\\\
expand the article with more detailed descriptions\\\\
find a related video to accompany the article
\end{tabular} & {\tt\begin{tabular}[t]{@{}p{\linewidth}@{}}
DescribeEnvironmentAndActivityBenefits(image\_patch)\\
YouTube\_v3.v3\_Search(part)\\
\colorbox{green!30}{TextExpander(text)}\\
inquire\_benefits\_of\_activity(activity, group)\\
play\_youtube\_video(video\_id)
\end{tabular}} \\
\bottomrule
\end{tabular}
\caption{Retrieval comparison for example taken from ToolRet's test set.}
\label{fig:qual4}
\end{figure*}
\endgroup

\begingroup
\setlength{\tabcolsep}{2.5pt}
\begin{figure*}[t!]
\footnotesize
\renewcommand{\arraystretch}{1.15}
\centering
\begin{tabular}{lP{6.4cm}P{8.1cm}}
\toprule
\rowcolor{gray!8}
\it User Query&\multicolumn{2}{P{14.5cm}}{I need resources for understanding "reinforcement learning" techniques, can you suggest some research papers and GitHub repositories?} \\
\it Targets &\multicolumn{2}{l}{\tt ResearchFinder, RepoTool} \\
\midrule
\bf Method & \bf Query & \bf Top-5 Retrieved \\
\midrule
\rowcolor{gray!8}
\ours & \begin{tabular}[t]{@{}p{\linewidth}@{}}
\tt <sub\_goal> Retrieve research papers related to a specific topic by processing the topic as input.</sub\_goal>\\
tool for retrieving research papers related to a specific topic, input topic string, output research papers, academic, paper database, paper search, literature retrieval\\
\tt <sub\_goal> Suggest GitHub repositories related to a specific topic by processing the topic as input. </sub\_goal> \\
tool for suggesting GitHub repositories related to a specific topic, input topic string, output GitHub repositories, repository search, code repository, GitHub API, research resources\\\\
tool for suggesting GitHub repositories related to a specific topic, input topic string, output GitHub repositories, GitHub API, repository search, code repository, research resources
\end{tabular} & {\tt\begin{tabular}[t]{@{}p{\linewidth}@{}}
sb3/dqn-Acrobot-v1\\
\colorbox{green!30}{RepoTool}\\
ppo-BreakoutNoFrameskip-v4\\
GitHub\\
\colorbox{green!30}{ResearchFinder}
\end{tabular}} \\
dense & I need resources for understanding "reinforcement learning" techniques, can you suggest some research papers and GitHub repositories? & {\tt\begin{tabular}[t]{@{}p{\linewidth}@{}}
ppo-BreakoutNoFrameskip-v4\\
ppo-PongNoFrameskip-v4\\
ppo-Pendulum-v1\\
sb3/ppo-CartPole-v1\\
sb3/dqn-MountainCar-v0
\end{tabular}} \\
\rowcolor{gray!8}
Q2E & research papers on reinforcement learning techniques, GitHub repositories for reinforcement learning, open-source reinforcement learnin
g projects, academic papers on RL algorithms, best reinforcement learning implementations on GitHub & {\tt\begin{tabular}[t]{@{}p{\linewidth}@{}}
ppo-PongNoFrameskip-v4\\
ppo-BreakoutNoFrameskip-v4\\
ppo-Pendulum-v1\\
sb3/ppo-CartPole-v1\\
sb3/dqn-Acrobot-v1
\end{tabular}} \\
Q2D & \tt \{"name": "research\_and\_code\_resources", "description": "A tool that suggests key research papers and GitHub repositories for understanding reinforcement learning techniques.”,  "parameters": \{"topic": \{"description": "The specific topic within reinforcement learning to focus on, such as 'Q-learning', 'deep reinforcement learning', 'policy gradients', etc.","type": "string"\}\}\} & {\tt\begin{tabular}[t]{@{}p{\linewidth}@{}}
ppo-Pendulum-v1\\
ppo-PongNoFrameskip-v4\\
ppo-BreakoutNoFrameskip-v4\\
sb3/ppo-CartPole-v1\\
sb3/dqn-Acrobot-v1
\end{tabular}} \\
\rowcolor{gray!8}
Q2P & Given a `research resource retrieval` task, retrieve tools that can provide research papers and GitHub repositories related to a specific topic, such as "reinforcement learning," by processing the topic input and returning relevant resources.
 & {\tt\begin{tabular}[t]{@{}p{\linewidth}@{}}
ppo-Pendulum-v1\\
ppo-PongNoFrameskip-v4\\
ppo-BreakoutNoFrameskip-v4\\
sb3/ppo-CartPole-v1\\
sb3/dqn-MountainCar-v0
 \end{tabular}} \\
Re-Invoke & \begin{tabular}[t]{@{}p{\linewidth}@{}}
find research papers on reinforcement learning techniques \\\\
find GitHub repositories related to reinforcement learning techniques
\end{tabular} & {\tt\begin{tabular}[t]{@{}p{\linewidth}@{}}
ppo-Pendulum-v1\\
ArxivSearch.get\_arxiv\_article\_information(query)\\
ppo-BreakoutNoFrameskip-v4\\
PaperAnalyzer\\
ppo-seals-CartPole-v0
\end{tabular}} \\
\bottomrule
\end{tabular}
\caption{Retrieval comparison for example taken from ToolRet's test set.}
\label{fig:qual5}
\end{figure*}
\endgroup

\begingroup
\setlength{\tabcolsep}{4pt}
\begin{table*}[ht!]
\small
\centering
\begin{tabular}{L{3.4cm} >{\columncolor[gray]{0.9}}c>{\columncolor[gray]{0.9}}c  cc cc cc >{\columncolor[gray]{0.9}}c>{\columncolor[gray]{0.9}}c}
\toprule
\multirow{3}{*}{\bf Method} 
    & \multicolumn{2}{c}{\bf In-Domain} 
    & \multicolumn{8}{c}{\bf Zero-Shot Transfer} \\
    
\cmidrule(lr){2-3}
\cmidrule(lr){4-11}
& \multicolumn{2}{c}{\cellcolor[gray]{0.9}\bf Avg} 
& \multicolumn{2}{c}{\bf Web*} 
& \multicolumn{2}{c}{\bf Code} 
& \multicolumn{2}{c}{\bf Custom} 
& \multicolumn{2}{c}{\cellcolor[gray]{0.9}\bf Macro-Avg} \\

\cmidrule(lr){2-3}
\cmidrule(lr){4-11}

& {N@10} & {C@10} 
& {N@10} & {C@10}
& {N@10} & {C@10}
& {N@10} & {C@10}
& {N@10} & {C@10}
\\
\midrule
\midrule
\texttt{bge-large} & 42.5 & 45.5 & 19.7&13.5 & 20.6 & 26.0&24.0 & 23.4 &21.4 & 21.0 \\
Q2E/ZS & 47.0 & 50.5 & 21.2 & 14.0   &  24.1& 31.1  & 28.5  & 27.8& 24.6&24.3\\
Q2P/SFT & 45.9& 49.1 & 17.6  & 13.0  & 22.5  & 29.0  & 34.7  & 35.9& 24.9&25.9\\
D2Q &47.1 &49.5  & 22.5 & 16.7  & 23.1 & 28.8 & 27.5 &25.8&24.3&23.8\\
Re-Invoke &52.1 &56.8 & 25.4 &19.6  & 25.0 & 32.5 & 34.2 &33.0&28.2&28.4\\
\ours-\textsc{Format} & \bf 56.7 & \bf 59.5 & 19.6 & 14.6  & 23.8  & 31.8  & 30.5  & 30.2 & 24.6& 25.5\\
\ours & 52.4 & 56.3 & \bf 21.9 & \bf 16.3 & \bf 26.6 &\bf 34.1 & \bf 38.0 & \bf 37.2 & \bf 28.9 & \bf 29.2 \\
\midrule
\texttt{ToolRet-bge-large} & 52.8 & 54.3 & 25.4 & 18.8 & 21.0 & 26.8 & 39.5 & 36.4 & 28.7  & 27.3 \\
Q2E/ZS & 54.1 & 56.4 &  25.6  &  19.0  & 24.7  & 29.9  & \bf 40.6  & 37.1& \bf 30.3& 28.7\\
Q2P/SFT &52.8 & 56.1 & 18.6  & 13.9 & 22.5  & 29.7  & 38.6 & 36.4& 26.6& 26.6\\
D2Q & \bf 58.7 & 61.6 & 28.0 & 18.6 & 27.6 & 33.9 & 38.2 & 34.8 & 31.3 & 29.1 \\
Re-Invoke & 58.4 & \bf 63.5 & \bf 28.3 & \bf 21.0 & \bf 28.3 & \bf 37.2 & 39.1 & 34.9 & 31.9 & 31.0 \\
\ours-\textsc{Format} & 57.7&  62.1 & 19.8 & 16.9 & 25.4  & 34.3  & 35.1  & 35.2  & 26.8  & 28.8\\
\ours & 56.4 & 60.8 & 21.8 & 17.4 &  28.0 &  36.0 & 40.5 & \bf 39.1 & 30.1 & \bf 30.8 \\
\midrule
\texttt{e5-mistral-7b} & 45.9 & 49.9 & 22.1&14.9 & 17.9& 24.3 &32.0 & 29.8 &24.0 & 23.0 \\
Q2E/ZS & 47.1& 51.2 & 22.1 & 15.3  & 21.0 & 28.3  & 35.7 & 35.6&26.3& 26.4\\
Q2P/SFT &50.8 & 54.5 & 23.4 & 17.4 & 29.3 & 38.1 & 41.3 & 37.5&31.3&31.0\\
D2Q & 52.9 & 55.9 & 23.8 & 16.2 & 27.0 & 33.9 & 29.2 & 27.0 & 26.7 & 25.7 \\
Re-Invoke & 54.5 & 60.1 & \bf 26.2 & 18.4 & 28.1 & 35.9 & 37.8 & 33.2 & 30.7 & 29.1 \\
\ours-\textsc{Format} & \bf 59.1 & \bf 63.3 & 21.5 & 17.8  &31.5  & 42.3  & 36.2 & 35.0&29.7&31.7\\
\ours & 55.6 & 60.5 & 24.4 & \bf 19.9 & \bf 33.8 & \bf 44.8 & \bf 43.3 & \bf 39.2 & \bf 33.9 & \bf 34.6 \\
\midrule
\texttt{e5-base} & 46.1 & 48.7 & 14.7 & 10.1 & 14.6 & 17.9 & 23.0 & 22.0 & 17.4 & 16.7 \\
Q2E/ZS & 47.0 & 49.8 & 15.4 & 10.9 & 17.3 & 21.4 & 24.6 & 24.7 & 19.1 & 19.0 \\
Q2P/SFT & 48.2 & 51.0 & 14.5 & 11.3 & 17.5 & 22.4 & 28.8 & 29.8 & 20.3 & 21.2 \\
D2Q & 48.4 & 50.3 & 19.7 & 13.8 & \bf 22.5 & \bf 28.3 & 23.7 & 23.0 & 22.0 & 21.7 \\
Re-Invoke & 51.4 & 56.4 & \bf 22.0 & \bf 16.1 & 21.4 & 27.3 & \bf 32.5 & 30.2 & \bf 25.3 & \bf 24.5 \\
\ours-\textsc{Format} & \bf 60.1 & \bf 60.6 & 13.9 & 12.7 & 18.0 & 23.2 & 24.0 & 25.9 & 18.6 & 20.6 \\
\ours & 52.2 & 55.1 & 17.0 & 13.2 & 21.5 & 27.5 & 31.9 & \bf 32.4 & 23.5 & 24.4 \\
\midrule
\texttt{ToolRet-e5-base} & 60.4 & 63.5 & 24.5 & 17.6 & 18.3 & 22.9 & 38.2 & 37.0 & 27.0 & 25.8 \\
Q2E/ZS & 60.5 & 63.6 & 25.3 & 19.3 & 22.5 & 30.1 & 39.4 & \bf 38.4 & 29.1 & 29.3 \\
Q2P/SFT & 57.2 & 60.7 & 22.1 & 16.4 & 20.2 & 27.9 & 37.0 & \bf 38.4 & 26.4 & 27.6 \\
D2Q & 57.6 & 61.2 & 25.2 & 17.8 & \bf 26.1 & 32.5 & 33.4 & 32.3 & 28.2 & 27.5 \\
Re-Invoke & 57.6 & 62.7 & \bf 25.6 & 18.5 & 25.1 & 32.0 & 35.1 & 33.0 & 28.6 & 27.8 \\
\ours-\textsc{Format} & \bf 61.3 & \bf 64.4 & 21.2 & 17.2 & 23.9 & 31.9 & 34.6 & 34.8 & 26.5 & 28.0 \\
\ours & 59.3 & 63.1 & 23.0 & \bf 19.5 & 25.2 & \bf 33.4 & \bf 39.9 & 38.0 & \bf 29.4 & \bf 30.3 \\
\bottomrule
\end{tabular}
\caption{Retriever transfer results on \textsc{ToolRet}. \ours is trained on \texttt{gte-qwen}-generated data, and used out-of-the-box directly with various retrievers at inference time. Web$^\star$ denotes the zero-shot Web datasets (excluding the in-domain training sources).}
\label{tab:ret_transfer}

\end{table*}
\endgroup
\begingroup
\setlength{\tabcolsep}{8.5pt}
\begin{figure*}[h]
\small
\centering
\begin{tabular}{l p{.89\linewidth}}
\toprule

\rowcolor{gray!12}
\bf Role & \bf Message \\
System & 
You are a Multi-Turn Tool Retrieval Planner, an expert AI assistant that deconstructs a user's query into a logical plan and then executes that plan to find the best tools.\newline\newline

You must perform one of the following:\newline
- To create the initial plan, you must output a {\tt <task\_breakdown>} block followed by a {\tt <sub\_goals>} block on the next line.\newline
- To execute your plan, you must output a single functional search query in a {\tt <query>}block.\newline
- To end the retrieval process, you must output the
 {\tt <stop\_retrieval>} tag when you have found tools for all sub-goals.\\
\rowcolor{gray!12}
User & User query: I'm planning a beach vacation with my friends and I need information about hotels in Cancun. Can you suggest some hotels in Cancun with a beachfront location, all-inclusive meal plan, and water sports facilities? Also, provide me with the hotel details and guest reviews. Additionally, I would like to know the cancellation probability of a flight from Los Angeles to Cancun on June 30, 2024, operated by American Airlines, flight number AA567. \\
Assistant& 
{\tt <task\_breakdown>}\newline
Given a `travel accommodation and flight information` task, retrieve tools that can suggest hotels based on location, amenities, meal plans, and guest reviews, as well as tools that provide the cancellation probability for a specific flight by processing parameters such as region, check-in/out dates, number of adults, desired amenities, flight details, and dates. The tools should return hotel suggestions with detailed information and reviews, as well as the cancellation probability for the specified flight.\newline
{\tt </task\_breakdown>}\newline
{\tt <sub\_goals>}\newline
{\tt
["Suggest hotels based on location, amenities, meal plans, and guest reviews.", "Retrieve hotel suggestions with detailed information and guest reviews based on location, amenities, meal plans, and check-in/out dates.", "Provide the cancellation probability for a specific flight based on flight details and dates."]}\newline
{\tt </sub\_goals>}\\
\rowcolor{gray!12}
User & Begin retrieval. \\
Assistant & 
hotels by filter domain string region\_id int check\_in date check\_out date adults int children int amenities array meal\_plan beachfront water\_sports all-inclusive accessibility boolean available only \\
\rowcolor{gray!12}
User & System retrieved tools for previous query: \{{\tt formatted\_retrieval\_results}\} \\
Assistant & 
hotel suggestion tool with detailed hotel information and guest reviews filters region date\_range amenities beachfront all-inclusive meal\_plan water\_sports \\
\rowcolor{gray!12}
User & System retrieved tools for previous query: \{{\tt formatted\_retrieval\_results}\} \\
Assistant & 
flight cancellation probability flight\_number string airline string origin string destination string flight\_date date \\
\rowcolor{gray!12}
User & System retrieved tools for previous query: \{{\tt formatted\_retrieval\_results}\} \\
Assistant & {\tt <stop\_retrieval>} \\
\bottomrule
\end{tabular}
\caption{Training (SFT) example for \ours.}
\label{fig:training_example}
\end{figure*}
\endgroup

\begin{figure*}[h]
\small
\centering
\begin{tabular}{L{.97\linewidth}}
\toprule
You are an expert AI assistant. Your task is to search for a specific set of 'Tools' that can be used to successfully solve the user's query. Given a user query, you should rewrite it into a search query for a dense retrieval system that covers all possible tools that can be used to solve this query. Your output should contain the rewritten search query only without any intermediate output.\\\\

User Query: \{{\tt query}\} \\
\bottomrule
\end{tabular}
\caption{Prompt for the Q2E/ZS baseline.}
\label{fig:prompt_q2e}
\end{figure*}
\begin{figure*}[h]
\small
\centering
\begin{tabular}{L{.97\linewidth}}
\toprule
You are an expert AI assistant. Your task is to search for a specific set of 'Tools' that can be used to successfully solve the user's query. Given a user query and possibly relevant context,
you should rewrite it into a search query for a dense retrieval system that covers all possible tools that can be used to solve this query. Your output should contain the rewritten search query only without any intermediate output.\\\\

Context:\\
\{{\tt prf\_context}\}\\
User Query: \{{\tt query}\} \\
\bottomrule
\end{tabular}
\caption{Prompt for the Q2E/PRF baseline.}
\label{fig:prompt_q2e_prf}
\end{figure*}
\begin{figure*}[h]
\small
\centering
\begin{tabular}{L{.97\linewidth}}
\toprule
You are an expert AI assistant. Your task is to create a 'Tool' that can be used to successfully solve the user's query. Given a user query, you should consider all tools that are needed to solve this query, and output one of them. Your output should contain a single tool in JSON format. An example tool is in this format:
{\tt \{"name": "tool name", "description": "tool description.", "parameters": \{"parameter a": \{"description": "parameter a description.", "type": "parameter a type"\}\}\}}\\\\

User Query: \{{\tt query}\} \\
\bottomrule
\end{tabular}
\caption{Prompt for the Q2D/ZS and HyDE/ZS baselines.}
\label{fig:prompt_q2d}
\end{figure*}
\begin{figure*}[h]
\small
\centering
\begin{tabular}{L{.97\linewidth}}
\toprule
You are an expert AI assistant. Your task is to create a 'Tool' that can be used to successfully solve the user's query. Given a user query and possibly relevant context, you should consider a
ll tools that are needed to solve this query, and output one of them. Your output should contain a single tool in JSON format.\\\\

Context:\\
\{{\tt prf\_context}\}\\
User Query: \{{\tt query}\} \\
\bottomrule
\end{tabular}
\caption{Prompt for the Q2D/PRF and HyDE/PRF baselines.}
\label{fig:prompt_q2d_prf}
\end{figure*}
\begin{figure*}[h]
\small
\centering
\begin{tabular}{L{.97\linewidth}}
\toprule
Suppose you are an assistant and you have access to the following API to answer user's queries. You are provided with a tool and its available API function including the description and parameters.\\\\

Your task is to generate a possible user query that can be handled by the API.\\\\

You must include the input parameters required in the API call. Please be creative and generate random but specific information for the required parameters.
Now you are given the API documentation below:\\\\

\{{\tt tool\_documentation}\}\\\\

Please generate a user query that you will need to call this tool. Note the generated query should be complex enough to describe the scenarios that you will need to call the provided API to address them.\\\\

The relevant query is:  \\
\bottomrule
\end{tabular}
\caption{Prompt for the D2Q/ZS and Re-Invoke baselines.}
\label{fig:prompt_d2q}
\end{figure*}
\begin{figure*}[h]
\small
\centering
\begin{tabular}{L{.97\linewidth}}
\toprule
**Instructions**\\
Suppose you are a query analyzer and your task is to extract the underlying user intents from the input query. You should preserve all the underlying user request and the extracted user intents should be easily understood without extra context information.
You should carefully read the given user query to understand its different intents. Then identify what are the specific intents. Each individual intent should be separated by a newline.\\\\

Here are some examples of how you should solve the task.\\\\

**Example**\\
Query: I'm planning to travel to Paris next weekend to visit my family, could you help me book a round trip flight ticket? I want to fly in economy class.\\
Intent:\\
book a round-trip flight ticket in economy class to Paris next weekend\\\\

Query: I'm a potential buyer looking for a condominium in the city of Miami. I am specifically interested in properties that have a minimum of two bathrooms. It should have walkable distance to the grocery stores.\\
Intent:\\
buy a real estate in Miami with a minimum of two bathrooms and walkable distance to the grocery stores\\\\

Query: I want to learn Spanish by talking to the native speakers at any time. Additionally, can you recommend some interesting books, preferably fictions, so that I can learn by reading? Also include the websites that I can buy them.\\
Intent:\\
learn Spanish by talking to the native speakers\\
recommend fictions to learn Spanish by reading\\
suggest the websites to buy Spanish fictions\\\\
**Begin!** \\
Query: \{{\tt query}\} \\
Intent:\\
\bottomrule
\end{tabular}
\caption{Prompt for the Re-Invoke baseline, for extracting user intents.}
\label{fig:prompt_reinvoke}
\end{figure*}
\begin{figure*}[h]
\small
\centering
\begin{tabular}{L{.97\linewidth}}
\toprule
\cellcolor[gray]{0.9}\it 1st Turn \\
\midrule
You are an expert AI Planner. Your goal is to find the best tools to solve a user's query by interacting with a tool retrieval system. You will be given the user's query.\\\\

For the first turn, you should deconstruct the user's query into a logical plan that you can follow to retrieve the correct tools by extracting the underlying user intents from the input query. You should preserve all the underlying user request and the extracted user intents should be easily understood without extra context information. You should carefully read the given user query to understand its different intents. Then identify what the specific intents are.\\\\

From the second turn on, you must generate a search query that targets an individual intent from the intents you extracted. Do not use any tags or formatting, any text you output will be used as the search query. A good tool query is a descriptive 'functional query' that captures the tool's purpose, possibly including likely parameter names (e.g., 'city: string'), categories, or library types to aid retrieval, but not specific filled-in parameter values (e.g. 'city: Chicago') that may harm retrieval. To end the retrieval process, you must output the\texttt{<stop\_retrieval>} tag without any other text when you have found the best tools for all intents.\\\\

User Query: \{{\tt query}\}. Breakdown of user's intents:\\
\midrule
\cellcolor[gray]{0.9}\it 2nd Turn \\
\midrule
Next search query (or \texttt{<stop\_retrieval>} if best tools found for all intents):\\
\midrule
\cellcolor[gray]{0.9}\it 3rd Turn+ (Retrieval Feedback) \\
\midrule
System retrieved tools for previous query:\\
\{{\tt formatted\_retrieved\_results}\}\\
\bottomrule
\end{tabular}
\caption{Prompt for the \ours-\textsc{Prompting} ablation study.}
\label{fig:prompt_toolqpp}
\end{figure*}

\begin{figure*}[h]
\small
\centering
\begin{tabular}{L{.97\linewidth}}
\toprule
\cellcolor[gray]{0.9}\it Sub-task Extraction Prompt \\
\midrule
You are an AI assistant that extracts key tasks. Given a high-level task breakdown and a specific target tool, extract the single, concise semantic goal from the breakdown that corresponds to that tool. Do not invent new goals that do not exist in the breakdown. Your output should be one single sentence or phrase, and not just keywords.

Task breakdown: \{{\tt natural\_language\_plan}\}.\\
Target tool name: \{{\tt tool\_name}\}.\\
\midrule
\cellcolor[gray]{0.9}\it Plan Alignment Prompt \\
\midrule
You are an expert planner. Your task is to analyze a user query, a task breakdown, and a list of available tools, then determine the correct sequential order to call the tools to solve the user's request.\\\\

**Instructions:**\\
1. Read the User Query and Task Breakdown to understand the user's multi-step goal.\\
2. Examine the 'Unordered Tools with Descriptions' to understand what each tool does.\\
3. Create a logical, step-by-step plan by ordering the provided tool names. The order should reflect the sequence of operations needed to fulfill the user's query from start to finish.\\
4. Your output MUST be a single, valid JSON list of tool names, and nothing else. The output list length must be equal to the number of tools in the given unordered list, with each tool appearing once only. \\\\

**[USER INPUT]**\\
User Query: \{{\tt query}\}\\
Task Breakdown: \{{\tt task\_breakdown}\}\\
Unordered Tools with Descriptions: \{{\tt tools}\}\\\\

**[YOUR CORRECT OUTPUT]**\\
\midrule
\cellcolor[gray]{0.9}\it Query Generation Prompt \\
\midrule
You are an expert AI Planner. Your goal is to find the best tools to solve a user's query by interacting with a tool retrieval system.
You will be given the user's query and an initial Task Breakdown that serves as your high-level plan.
For each step, you must generate a search query in a \texttt{<query>} tag. Do not use any other tags such as \texttt{<tool\_call>} or \texttt{<reasoning>} or \texttt{<think>}.
A good tool query is a descriptive 'functional query' that captures the tool's purpose, possibly including likely parameter names (e.g., 'city: string'), categories, or library types to aid retrieval, but not specific filled-in parameter values (e.g. 'city: Chicago') that may harm retrieval.\\\\

\relax[Overall User Goal]: \{{\tt query}\}\\
\relax[Your Overall Plan]: \{{\tt natural\_language\_plan}\}\\
\relax[Previously Completed Steps]:\\
\relax[Step \{{\tt i}\}]\\
Goal: "\{{\tt goal}\}"\\
Successful Query: "\{{\tt search\_query}\}"\\
... \\
--- \\
\{{\tt tool\_context}\} \\
--- \\
\relax[Instruction]: Write the functional query within a <query></query> tag (e.g., `<query>your functional query</query>`).\\
\midrule
\cellcolor[gray]{0.9}\it AddMoreInfo (1st attempt) \\
\midrule
\relax[Goal For Your Current Step]: Find a tool for "\{{\tt semantic\_goal}\}"\\
Task to focus on: craft a functional query to retrieve a tool that can address the current step goal. You can include information or keywords from the goal in your functional query\\
\midrule
\cellcolor[gray]{0.9}\it AddMoreInfo (2nd attempt) \\
\midrule
We should look for a tool with this description: "\{{\tt tool\_description}\}" \\
\midrule
\cellcolor[gray]{0.9}\it AddMoreInfo (3rd attempt) \\
\midrule
Previously retrieved with query "\{{\tt search\_query}\}", here's the retrieved tools:\\
\{{\tt retrieval\_results}\}\\
\midrule
\cellcolor[gray]{0.9}\it AddMoreInfo (4th attempt) \\
\midrule
Previously retrieved with query "\{{\tt search\_query}\}", tools: \{{\tt retrieval\_results}\}\\
Previously retrieved with query "\{{\tt search\_query+natural\_language\_plan}\}", tools: \{{\tt retrieval\_results}\}\\
\midrule
\cellcolor[gray]{0.9}\it AddMoreInfo (5th attempt) \\
\midrule
Hint: Full Target Tool Info: \{{\tt json.dumps(tool\_documentation\_without\_toolname)}\}.\\
Include information from 'description' and relevant parameter names in the functional query.\\
\bottomrule
\end{tabular}
\caption{Prompts for data generation for training \ours.}
\label{fig:prompt_toolqp}
\end{figure*}

\end{document}